\documentclass[conference]{IEEEtran}
\usepackage[T1]{fontenc}
\usepackage{amsmath,amssymb}
\usepackage{graphicx}
\usepackage{booktabs,tabularx,array}
\usepackage{cite}
\usepackage[hidelinks]{hyperref}
\usepackage{microtype}
\newcommand{\TruthCases}{81}
\newcommand{\TruthPassed}{81}
\newcommand{\MutationCases}{192}
\newcommand{\MutationRejected}{192}

\newcommand{\EndpointMedian}{0.49}
\newcommand{\CheckerMedian}{51.20}
\newcommand{\BaselineRows}{%
Complete-case reference & 0 & 0 & 65 \\
\midrule
Support-only & 1 & 65 & 0 \\
Relationship-only & 7 & 38 & 27 \\
Disclosure-only & 7 & 38 & 27 \\
\midrule
Support-only + guard & 1 & 0 & 65 \\
Relationship-only + guard & 7 & 0 & 65 \\
Disclosure-only + guard & 7 & 0 & 65 \\
\midrule
Drop $I$ + guard & 1 & 0 & 65 \\
Drop $A$ + guard & 1 & 0 & 65 \\
Drop $M$ + guard & 1 & 0 & 65 \\
Drop $D$ + guard & 1 & 0 & 65 \\
Drop $d_Q$ (priority only) & 0 & 0 & 65 \\
}
\newcommand{\PropertyRows}{%
Complete truth table & 81/81 \\
Evidence removal & 216/216 \\
Support separation & 81/81 \\
Query-priority separation & 891/891 \\
Controlled binary pairs & 32/32 \\
Canonical round trip & 81/81 \\
Annotation-version lineage & 81/81 \\
Reject malformed records & 192/192 \\
}

\hypersetup{pdftitle={Claim-Gated Source-Risk Auditing for Generative Search},pdfauthor={Kainan Zhou; Chuhong Xu; Gangzhen Qian; Zhaoyi Li},pdfsubject={Source-risk audit specification and synthetic contract-conformance validation}}
\newcommand{\U}{\mathrm{U}}
\newcommand{\RO}{R_{\mathrm O}}
\newcommand{\Tpost}{T_{\mathrm{post}}}
\newcommand{\authorcell}[4]{\begin{minipage}[t]{0.245\textwidth}\centering
{\normalsize #1}\par\vspace{2pt}{\fontsize{8.5}{10}\selectfont #2\par #3\par #4}\end{minipage}}
\title{Claim-Gated Source-Risk Auditing\\for Generative Search}
\author{\authorcell{Kainan Zhou}{Google LLC}{Mountain View, USA}{zhoumark@google.com}\hfill
\authorcell{Chuhong Xu}{Sony Corporate of America}{San Jose, USA}{chuhong.xu@sony.com}\hfill
\authorcell{Gangzhen Qian}{Google LLC}{Mountain View, USA}{irisqian@google.com}\hfill
\authorcell{Zhaoyi Li}{Intuit Inc}{Mountain View, USA}{lzy9776@gmail.com}}
\begin{document}
\maketitle
\begin{abstract}
A generative search answer can cite a supported passage yet omit a source relationship that changes its interpretation. We specify a claim-gated audit of the query--source--answer tuple. An omission is resolved only when relationship evidence, answer adoption, materiality, and disclosure are all observed; incomplete evidence remains unresolved rather than being treated as independence. The specification separates this endpoint from citation support and review priority, and binds decisions to versioned evidence spans. A reference checker makes the record contract executable. On an exhaustive synthetic suite, it reproduces all \TruthCases{} three-state predicate combinations and rejects \MutationCases{} deliberately malformed records. Common-guard baselines and predicate ablations isolate endpoint logic from missing-evidence handling, while controlled transitions check support separation and evidence removal. These are finite contract-conformance results, not detector accuracy or evidence of improved user outcomes. We define the independent annotation, held-out evaluation, and paired utility tests still required to establish semantic validity and deployment benefit.
\end{abstract}

\section{Introduction}
Generative search combines retrieval, source selection, attribution, and generation in a single answer. Existing verifiability audits ask whether citations support the statements attached to them \cite{liu2023}. A different question arises when a supported statement comes from a source with a relationship to the queried target. The statement may be accurate, but interpreting it as an independent assessment may still be unwarranted.

Consider an illustrative answer that recommends a product using a review written by a paid affiliate. If the answer adopts the recommendation without identifying the relationship, citation support does not resolve the missing context. Conversely, a company-authored specification can be used appropriately when its origin is clear and the answer makes a factual comparison. Neither ownership nor commercial status alone establishes an omission. The relevant unit must include the query, the source, and the answer's treatment of that source.

Citation-generation benchmarks measure correctness and citation quality \cite{gao2023}. Publisher-side optimization of answer inclusion provides a further reason to retain source context \cite{aggarwal2024}, but visibility or persuasive language is not evidence of a particular relationship. The audit therefore requires recoverable relationship evidence and a separate judgment of whether its omission matters for the adopted claim.

This paper contributes an executable specification rather than a new neural architecture. Its first contribution is a four-predicate omission endpoint with an explicit complete-case policy. Its second is a record contract that separates source evidence, answer-level judgments, and post hoc review priority. Its third is a reproducible conformance suite with common-guard baselines, component ablations, and malformed-record tests. These components make a narrower claim than a deployed detector: given explicit judgments, the implementation preserves their meaning and reports only the validation stage actually completed.

The distinction matters for evaluation. An implementation can reproduce a formula without recognizing relationships in Web text, and a valid evidence pointer can lead to an incorrect human judgment. We test the former, not the latter. The resulting artifact is a basis for subsequent semantic evaluation; it does not estimate omission prevalence, demonstrate better answers, or certify publisher trustworthiness.

\section{Related Work and Scope}
\subsection{Citation Support and RAG Evaluation}
RAGTruth supplies a hallucination corpus for retrieval-augmented generation \cite{niu2024}, while RAGAs evaluates retrieval and generation components \cite{es2024}. Their content-level measurements are relevant to factual support but do not by themselves supply the archived source--target relationship required here. Support is retained as a separate field so that factual and contextual failures can be inspected together without being conflated.

ARES uses learned evaluators for retrieval-augmented systems \cite{saad2024}. Such evaluators could help identify candidate tuples or assist an annotator, but their predictions would remain distinct from relationship evidence. Our reference checker does not run these systems or compare their published scores: its inputs are structured judgments, whereas their evaluation tasks involve generated text and retrieved passages.

\subsection{Relationship Evidence and Disclosure}
Research on affiliate abuse traces relationships through transaction context \cite{chachra2015}. AdIntuition studies disclosure of likely endorsements on video pages \cite{swart2020}. These settings motivate a separation between finding a relationship and deciding whether the audited answer explains it adequately. A disclosure keyword elsewhere on a page is not sufficient evidence that the answer preserves the relevant context.

\subsection{Behavioral Tests and the Claim Boundary}
CheckList uses targeted behavioral tests to expose failures that an aggregate metric can miss \cite{ribeiro2020}. We adopt that testing principle at the structured-record level: change one field, register the expected endpoint transition, and preserve the rest. Unlike a natural-language benchmark, our exhaustive state space has specification-derived outcomes. Agreement with those outcomes establishes implementation conformance only; it cannot establish that the predicates are meaningful or reliably recoverable from natural text.

\section{Audit Specification and Reference Checker}
\subsection{Unit, Predicates, and Complete-Case Endpoint}
Let $r=(q,u,a)$ denote a query, an archived source available to the generator or cited in its output, and the resulting answer. Multiple sources produce separate tuples. A versioned codebook assigns four predicates in $\{0,1,\U\}$: $I$ records a query-relevant relationship; $A$ records adoption of the source's claim; $M$ records whether the relationship affects interpretation; and $D$ records adequate disclosure in the answer. Here $\U$ means that the available record does not resolve the judgment, not that the relationship is absent.

We use the complete-case endpoint
\begin{equation}
\RO(r)=
\begin{cases}
\U, & \text{if any of } I,A,M,D \text{ is }\U,\\
I A M(1-D), & \text{otherwise}.
\end{cases}
\label{eq:endpoint}
\end{equation}
Thus $(1,1,1,0)$ is an omission and $(1,1,1,1)$ is not. The tuple $(0,1,1,\U)$ remains unresolved even though short-circuit Boolean reasoning could return zero. This is a deliberate documentation requirement: a resolved record must contain all four judgments. It is stricter than merely determining whether the conjunction can be true.

The endpoint concerns omission of material context in the observed answer. It does not attribute intent, identify deception, or establish a causal effect of the source on the generator. Adoption and materiality require their own evidence; neither follows automatically from a relationship label.

\subsection{Evidence Vector and Review Priority}
The record retains five dimensions,
\begin{equation}
\mathbf d(r)=(d_Q,d_I,d_A,d_M,d_G)
\in \bigl([0,1]\cup\{\U\}\bigr)^5.
\label{eq:vector}
\end{equation}
They represent query susceptibility, relationship evidence, adoption, materiality, and disclosure deficit. In the reference checker, resolved entries map explicitly as $d_I=I$, $d_A=A$, $d_M=M$, and $d_G=1-D$; an unresolved predicate maps to $\U$. These entries are not class probabilities and need not sum to one. A later graded codebook would require a separately validated mapping to the predicates.

When all five dimensions are observed, review priority is
\begin{equation}
\Tpost(r)=\sum_k w_k d_k,
\quad w_k\geq0,\quad \sum_k w_k=1.
\label{eq:priority}
\end{equation}
This score orders review; it never changes (\ref{eq:endpoint}). The checker returns an unresolved priority when any dimension is missing, even when its weight is zero. The experiments use equal weights as an illustrative configuration, not as an optimized policy. Query susceptibility can support sampling, but answer-dependent dimensions cannot select the evidence used to generate that same answer.

\begin{figure}[t]
\centering
\includegraphics[width=\columnwidth]{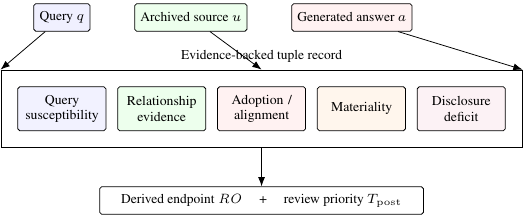}
\caption{Evidence flow for one tuple. The relationship, adoption, materiality, and disclosure fields resolve $\RO$; $d_Q$ affects review priority only. The reference checker consumes explicit judgments, not raw-text predictions.}
\label{fig:workflow}
\end{figure}

\subsection{Module Interfaces and Execution Order}
Figure~\ref{fig:workflow} separates the inputs and outputs. An archive adapter stores the query, source and answer snapshots, and pipeline configuration. An annotation adapter supplies the codebook version, predicate values, and evidence spans. The contract checker verifies representation and traceability. Only a conforming record reaches the endpoint resolver and priority scorer; their outputs then enter a claim ledger.

The execution order is deterministic: validate required fields and hashes; validate predicate types and evidence offsets; verify the dimension mapping and weights; resolve $\RO$; compute priority; and attach the completed validation gate. A malformed record is rejected with a reason rather than converted into a semantic label. A structurally valid record with unresolved judgments is accepted and returns $\U$. This distinction prevents parsing failures from silently becoming negative findings.

The implemented adapter generates synthetic records and supplies their judgments directly. Web collection, entity resolution, natural-language annotation, and generation remain external interfaces, not implemented capabilities of this checker. An optional learned annotator would need to be frozen and evaluated against independent labels before any accuracy claim.

\subsection{Source Roles and Admissible Evidence}
Table~\ref{tab:roles} retains the operational source-role taxonomy. Roles determine what evidence to seek and what disclosure the codebook should require. They are not standalone risk scores. Current ownership evidence may resolve $I$, but $A$, $M$, and $D$ still concern the particular answer.
\begin{table}[t]
\caption{Operational Source-Role Taxonomy}
\label{tab:roles}
\centering\footnotesize
\begin{tabularx}{\columnwidth}{@{}p{0.31\columnwidth}X@{}}
\toprule
Role & Operational interpretation \\
\midrule
Target owner & Observable ownership of the target-related source.\\
Target affiliate & Evidenced related party promoting or comparing the target.\\
Sponsored third party & An evidenced material sponsorship relationship.\\
Advocacy aligned & Observable organizational or cause alignment.\\
Commercial third party & Commercial publisher; relationship subtype unspecified.\\
Apparently independent & No visible ownership claim, but an archived material tie.\\
No observed relation & Prescribed search found no tie; independence unverified.\\
Community / UGC & Identity or coordination remains uncertain.\\
\bottomrule
\end{tabularx}
\end{table}

Evidence must be recoverable from an archived observation: for example, ownership records, sponsorship disclosures, affiliate records, or reproducible identity metadata. Tone, rank, similarity, and commercial status may guide investigation but do not resolve a relationship. Each decision is tied to the source version and time window used for the answer. Later evidence creates a new record version rather than rewriting the earlier observation.

\subsection{Record Contract and Invariants}
The contract stores query and tuple identifiers, original and canonical URLs, rank, collection time, source and answer snapshots, pipeline metadata, source role, codebook version, predicate judgments, and evidence offsets. Every resolved judgment has a recoverable span; unresolved synthetic judgments use JSON \texttt{null}. Missing keys, the string \texttt{"U"}, and numeric zero are not interchangeable.

The reference implementation derives snapshot digests with SHA-256. A tuple digest binds the query identity, source and answer digests, and pipeline configuration; a record digest additionally binds annotations and other stored fields. An annotation revision retains the tuple identity, changes the record identity, and points to its parent. Hashes support change detection, not authenticity or semantic correctness. Append-only storage enforcement remains a deployment responsibility.

Six invariants constrain the design: citation support is separate from source risk; findings remain tuple-local; decisions retain evidence pointers; missing evidence cannot create a resolved low-risk conclusion; generation and post hoc review remain separate; and stronger empirical claims require stronger validation. The conformance suite tests the specified record transitions, not every possible implementation of these principles.

\section{Executable Conformance Evaluation}
\subsection{Questions, Fixtures, and Oracle}
The evaluation asks whether the checker implements the declared endpoint, whether simpler rules preserve that contract, and whether the record validator rejects selected structural defects. We enumerate the complete state space $\{0,1,\U\}^4$: \TruthCases{} tuples comprising one positive, 15 negatives, and 65 unresolved cases under (\ref{eq:endpoint}). Each tuple has a synthetic query, source and answer snapshot, version metadata, and explicit judgment spans. The text contains fixture annotations, not naturally occurring Web claims.

There is no training, preprocessing of a real dataset, or train/test split. Enumerating a finite contract requires neither model fitting nor statistical sampling. Expected outputs are stored separately using membership in the explicit positive set $\{(1,1,1,0)\}$ for fully observed states and $\U$ otherwise. This is an independently expressed test oracle for the same rule, not independent semantic ground truth. The artifact contains no human labels and no live-Web observations.

All experiments run in Python 3.13.5 using only the standard library. The supplied scripts generate JSONL fixtures, the expected state table, per-case predictions, mutation logs, CSV summaries, and a checksum manifest. The paper's experimental tables are generated from these outputs. No paid API, pretrained model, or network access is required.

\subsection{Baselines and Common-Guard Comparison}
We compare three explicit rule projections. Support-only returns zero when passage support is present; all main fixtures have support fixed to one. Relationship-only returns $I$. Disclosure-only returns $1-D$ when $D$ is observed and $\U$ otherwise. The last rule is an oracle-feature proxy for a disclosure-only policy, not an executed keyword detector. All methods consume the same structured fixture judgments, so the comparison measures information discarded by each rule rather than text-understanding performance.

Table~\ref{tab:baselines} reports mismatches on the 16 resolved contract states, resolutions issued on the 65 states that the contract requires to remain unresolved, and the total number of unresolved outputs. We then apply the identical four-predicate completeness guard to all three baselines. This controls for missingness policy instead of crediting the reference method merely for having a stricter guard.

\begin{table}[t]
\caption{Contract Comparisons on the Exhaustive State Space}
\label{tab:baselines}
\centering\footnotesize
\setlength{\tabcolsep}{3pt}
\begin{tabularx}{\columnwidth}{@{}Xrrr@{}}
\toprule
Method & $E_{16}$ & $F_{65}$ & $U_{81}$\\
\midrule
\BaselineRows
\bottomrule
\end{tabularx}
\vspace{3pt}
\parbox{\columnwidth}{\scriptsize $E_{16}$: mismatches on 16 resolved contract states. $F_{65}$: resolutions contrary to the complete-case contract on 65 incomplete states. $U_{81}$: unresolved outputs among all 81 states. These are specification counts, not estimated detector error rates.}
\end{table}

Without the guard, support-only misses the sole positive state and resolves all 65 incomplete states. Relationship-only and disclosure-only each disagree on seven resolved states and resolve 38 incomplete states. With the common guard, those unsupported resolutions disappear, but the resolved-state mismatches remain. A guarded support-only rule agrees on 15 of 16 complete states simply because the positive state is rare in this enumeration. Its apparently high agreement therefore does not show that support can replace relationship auditing.

\subsection{Predicate Ablations and Coverage Trade-off}
Each endpoint ablation preserves the same completeness guard and removes one Boolean requirement: dropping $I$, $A$, or $M$ substitutes one; dropping $D$ substitutes zero. Each introduces one additional positive on the 16 fully observed states. For example, dropping $A$ incorrectly treats $(1,0,1,0)$ as an omission although the answer does not adopt the source claim. Removing $d_Q$ from the endpoint interface changes no outcome, as intended; this is a separation control rather than evidence that query susceptibility is useful for triage.

A legitimate alternative is to resolve a negative as soon as $I=0$, $A=0$, $M=0$, or $D=1$, even if another predicate is unknown. That short-circuit policy resolves 66 of 81 states, compared with 16 for our complete-case contract. Its 50 additional resolved negatives are not Boolean errors. They expose the cost of requiring a fully documented audit record. These synthetic coverage counts are not deployment rates, and the stricter policy is not claimed to be universally preferable.

\subsection{Controlled Transitions and Structural Mutations}
Table~\ref{tab:checks} summarizes the completed checks. The resolver matches all \TruthPassed{}/\TruthCases{} oracle outputs. Replacing one observed predicate with $\U$ produces 216 directed evidence-removal transitions; every destination remains unresolved. Toggling passage support leaves all 81 endpoints unchanged. Sweeping $d_Q$ from zero to one in steps of 0.1 yields 891 checks without changing an endpoint. These counts share base states and must not be treated as independent observations.

\begin{table}[t]
\caption{Finite Conformance Checks, Not Semantic Accuracy}
\label{tab:checks}
\centering\footnotesize
\begin{tabularx}{\columnwidth}{@{}Xr@{}}
\toprule
Check family & Passed / tested\\
\midrule
\PropertyRows
\bottomrule
\end{tabularx}
\end{table}

The binary suite flips one predicate while fixing the other three, giving 32 controlled pairs. For disclosure, changing $D$ from zero to one removes the omission only when $I=A=M=1$; the other seven pairs remain negative. The equivalent relationship, adoption, and materiality pairs test their registered directions. These are record-level interventions with supplied judgments, not paired generated answers evaluated by users.

All 81 valid records survive canonical serialization and a version-lineage check. The mutation suite applies 12 defect types to each of the 16 fully observed records, producing \MutationCases{} malformed cases. Defects cover identity, snapshot and record checksums, missing codebook or pipeline fields, invalid nullable labels, inconsistent dimensions, missing or invalid spans, mismatched quotes, and unnormalized weights. All \MutationRejected{} are rejected. Except for deliberate checksum or identity defects, records are resealed after mutation so that a stale outer digest cannot mask the intended structural test. This curated suite is not arbitrary-input fuzzing or a security proof.

\subsection{Measured Checker Cost and Interpretation}
A single-process microbenchmark on an Intel Xeon Platinum 8573C host records 21 repeated batch means after warm-up. The endpoint loop processes 16,200 tuples per repeat; the full record checker processes 810. The median batch-mean cost is \EndpointMedian{}~$\mu$s per tuple for the endpoint and \CheckerMedian{}~$\mu$s for validation plus auditing. These measurements exclude archive I/O, retrieval, generation, and annotation; they are neither end-to-end latency nor per-request tail-latency estimates.

The experiment now supplies an executable record checker and numerical comparisons, but its success criterion is agreement with a chosen specification. An incorrect materiality label can still pass every structural test. No confidence interval or significance test is attached to the exhaustive counts, and repeated template cases are not used to claim a large empirical sample. The completed claim is finite conformance of the supplied implementation.

\section{Semantic Validation and Application Boundary}
\subsection{Independent Labels and Held-Out Evaluation}
\begin{figure}[t]
\centering
\includegraphics[width=\columnwidth]{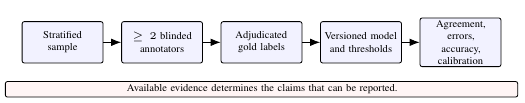}
\caption{Required semantic validation sequence. These human-label and model-evaluation stages remain future work; they are not established by the synthetic conformance checks.}
\label{fig:validation}
\end{figure}

Figure~\ref{fig:validation} shows the next validation stage, which has not been completed. A Web study must archive observations before annotation, stratify by application domain and prespecified risk bands, and record inclusion probabilities. Target families, related domains, and query templates should not leak from calibration into the held-out partition. Synthetic fixtures remain outside the human-gold test set.

At least two trained annotators should label each tuple independently, blinded to model scores and one another's decisions. They must be able to assign $\U$ and cite evidence for resolved predicates. Adjudication follows, while reliability is reported on the original independent labels using a coefficient appropriate to the label scale \cite{artstein2008}. The codebook, pipeline, and operating thresholds must be frozen before opening the test labels.

The primary semantic measurements should separate each predicate from the derived endpoint. Confusion matrices, class-specific counts, unresolved coverage, and domain-clustered intervals expose errors hidden by aggregate classification measures \cite{sokolova2009}. A learned judge should not become its own gold standard: work on LLM-based evaluation also reports potential evaluator biases \cite{geval2023}. Any judge-assisted annotation would therefore need a separately labeled audit subset and explicit reporting of assistance.

\subsection{Tool Utility and Deployment Decisions}
Application benefit requires a matched comparison of generated answers before and after an audit-informed intervention, with generation settings held fixed and evaluators blinded to condition. The primary endpoint is material relationship omission; utility measures include answerability, retained evidence, factual support, abstention, and end-to-end latency. Merely suppressing answers or removing related sources does not establish a useful defense.

A policy may return PASS, CONTEXTUALIZE, REVIEW, or ABSTAIN. PASS means that the configured review requirement is met for the observed tuple, not that the publisher is independent or universally reliable. CONTEXTUALIZE can request adequate relationship attribution while retaining useful evidence. REVIEW preserves unresolved cases; ABSTAIN requires an explicit answerability or evidence policy. These actions are proposed interfaces, not validated production controls.

\subsection{Limitations, Falsifiability, and Release}
Table~\ref{tab:gates} makes the remaining claim boundaries explicit. The present artifact validates typed synthetic records and deterministic logic. It does not establish source discovery, temporal validity of real relationships, annotation reliability, multilingual transfer, or changes in reader interpretation. Materiality and adequate disclosure remain task-dependent judgments. A hash-valid record can preserve an erroneous or outdated judgment perfectly.

\begin{table}[t]
\caption{Claim Gates and Evidence Status}
\label{tab:gates}
\centering\footnotesize
\begin{tabularx}{\columnwidth}{@{}p{0.25\columnwidth}Xp{0.16\columnwidth}@{}}
\toprule
Claim & Required evidence & Status\\
\midrule
Finite conformance & Executable oracle, fixture predictions, mutation logs & Completed\\
Real-Web finding & Archived observations, timestamps, retrieval provenance & Not tested\\
Semantic validity & Independent labels, reliability, adjudication & Not tested\\
Detector accuracy & Frozen system, held-out labels, class metrics and intervals & Not tested\\
Application benefit & Matched interventions, omission and utility measures & Not tested\\
\bottomrule
\end{tabularx}
\end{table}

Several findings would count against the specification: annotators cannot apply $M$ consistently; the role taxonomy systematically excludes relevant ties; a simpler provenance rule performs equally well against independent labels; the complete-case requirement causes impractical review load; or contextualization harms answer utility without reducing omissions. Any such result requires revising the affected rule rather than treating conformance as proof of usefulness.

The accompanying package includes the checker, fixture generator, test oracle, per-case outputs, aggregate CSVs, and table-generation script. A checksum manifest binds the files used for each run; it is not a digital signature.\newpage
\vspace*{18pt}
The future Web release must additionally document archive permissions, exclusions, codebook changes, adjudication, and split assignments. Restricted evidence can retain authorized pointers, but redaction must not silently change the reported sample.

\section{Conclusion}
Source-risk auditing asks whether a generated answer preserves a material relationship, not only whether a citation supports its words. We formalize that question at the query--source--answer level and implement a record checker with explicit missingness, evidence pointers, and separate review priority. Exhaustive predicate enumeration, common-guard comparisons, component ablations, and structural mutations establish finite conformance of the implementation. They also expose the coverage cost of a complete-case policy. Independent human labels and paired answer-level evaluations remain necessary before claiming semantic accuracy, omission prevalence, or application benefit.

\section*{Acknowledgment}
We acknowledge that this work was supported by Beijing Institute of Technology, Zhuhai (Project No. 2026039DCXM).

\end{document}